\documentclass[a4paper]{article}
\usepackage[latin9]{inputenc}
\usepackage{array}
\usepackage{float}
\usepackage{multirow}
\usepackage{amsmath}
\usepackage{amssymb}
\usepackage{graphicx}

\makeatletter

\pdfpageheight\paperheight
\pdfpagewidth\paperwidth

\providecommand{\tabularnewline}{\\}

\usepackage{INTERSPEECH2015}

\usepackage{bm}

\ninept

\title{Residual Memory Networks: Feed-forward approach to learn \\ long temporal dependencies}
\name{\parbox{0.9\linewidth}{\center 
 Murali Karthick Baskar, Martin Karafi\'{a}t, Luk\'{a}\v{s} Burget, Karel Vesel\'{y}, Franti\v{s}ek
  Gr\'{e}zl \linebreak  and Jan ``Honza'' \v{C}ernock\'{y}}
}
\address{
Brno University of Technology, Speech@FIT and IT4I Center of
  Excellence, Brno, Czech Republic\\
  {\small \tt \{baskar,karafiat,burget,iveselyk,grezl, cernocky\}@fit.vutbr.cz}
}

\makeatother

\begin{document}
\ninept\maketitle 
\begin{abstract}
Training deep recurrent neural network (RNN) architectures is complicated
due to the increased network complexity. This disrupts the learning
of higher order abstracts using deep RNN. In case of feed-forward
networks training deep structures is simple and faster while learning
long-term temporal information is not possible. In this paper we
propose a residual memory neural network (RMN) architecture to model
short-time dependencies using deep feed-forward layers having residual
and time delayed connections. The residual connection paves way to
construct deeper networks by enabling unhindered flow of gradients
and the time delay units capture temporal information with shared
weights. The number of layers in RMN signifies both the hierarchical
processing depth and temporal depth. The computational complexity
in training RMN is significantly less when compared to deep recurrent
networks. RMN is further extended as bi-directional RMN (BRMN) to
capture both past and future information. Experimental analysis is
done on AMI corpus to substantiate the capability of RMN in learning
long-term information and hierarchical information. Recognition performance
of RMN trained with 300 hours of Switchboard corpus is compared with
various state-of-the-art LVCSR systems. The results indicate that
RMN and BRMN gains 6 \% and 3.8 \% relative improvement over LSTM
and BLSTM networks.
\end{abstract}
\noindent \textbf{Index Terms}: Automatic speech recognition, LSTM,
RNN, Residual memory networks.

\section{Introduction}

Automatic speech recognition (ASR) has largely improved using recurrent
neural network (RNN) acoustic models due to the networks ability to
learn long-term information. Unfortunately, RNNs becomes difficult
to train when extended to deeper structure. This is  because the deeper
models is essential to learn more abstract information for improving
the prediction of unseen data \cite{sak2014long}. Several attempts
have been made to train deep RNN such as using non-recurrent structures
to increase the number of layers and model complexity \cite{pascanu2013construct},
including transform gates for smoother gradient flow \cite{zhang2016highway}.
But training deep recurrent structures is complex as the gradient
has to travel multiple hidden states and lack of better optimization
algorithms \cite{pascanu2013construct}. Meanwhile, deep neural networks
(DNN) can run much deeper, lead to better generalization to unseen
data and are less prone to overfitting. Also, feed-forward training
is relatively simple and faster when compared to recurrent structures.
Besides these advantages, DNN fail to perform better for tasks which
require long-term information. Thus to overcome this constraint, the
authors in \cite{zhang2015feedforward} represented the temporal context
as a fixed size representation and train them jointly. In another
work by \cite{raffel2015feed,vesel2016sequence}, the unweighted average
of input context is used in modeling temporal information. The limitation
with these approaches are that they fail to model the temporal order
which is important for speech tasks as shown in \cite{hochreiter1997long}.
Analogous to these networks is time delay neural network (TDNN), where
each layer is fed with multi spliced input \cite{peddinti2015time}.
Even though TDNN has the ability to model long-term contexts, RNN
still shows better performance over TDNN \cite{peddinti2015time}.
A possible reason for this is the fact that the performance gain of
RNN is devoted to their stepwise learning of time frames and not by
the size of context as found in \cite{mohamed2015deep}.

\begin{figure}[H]
\begin{centering}
\includegraphics[scale=0.4]{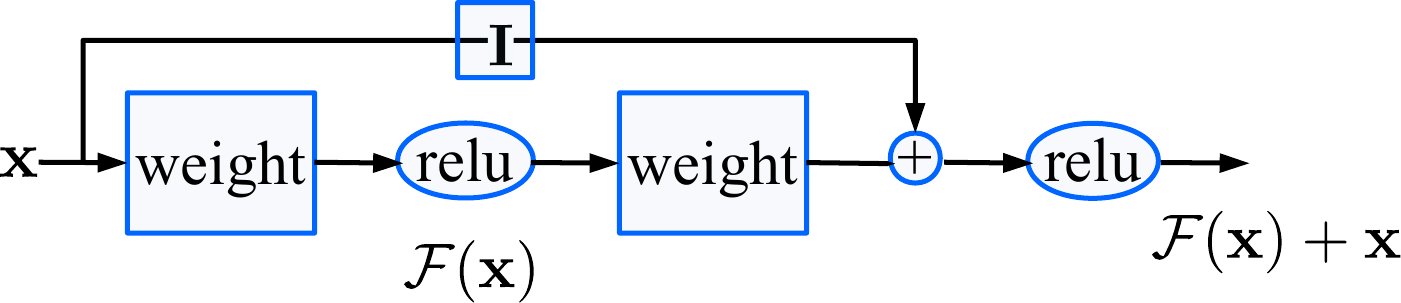}
\par\end{centering}

\caption{\label{fig:residual}Structure of residual component }
\end{figure}

A straight forward approach to allow DNN to learn a single time step
at each layer is by denoting the time context length based on the
number of layers. A practical challenge in this approach is that DNN
having more than few layers starts to degrade due to gradient vanishing
problem. Recent works by \cite{he2015deep} showed that deeper convolutional
networks can be trained in a much simpler way using residual connections.
This approach showed significant gains in image recognition tasks.
Additionally, \cite{ghahremani2016linearly} suggested to use shortcut
connections and singular value decomposition for deeper fully connected
networks and got better performance in ASR tasks. This paper makes
an attempt to use residual structure with time delayed connections
to harness the power of both temporal and hierarchical structures
at each layer as explained in section \ref{sec:Residual-memory-networks}.
Figure \ref{fig:residual}, illustrates an basic form of residual
network structure where the input $x$ is summed to next layers output
using a shortcut connection with identity mapping $I$.\vspace{-0.4cm}
\vspace{-0.4cm}
In this work, we propose Residual memory network (RMN), a variant
of DNN where the number of layers denotes both the length of temporal
information learnt and the structural depth. The key contributions
of RMN are:
\begin{itemize}
\item The use of residual connections after every few layers to increase
the network depth makes training faster with increase in performance.
During backpropagation, the residual lines allows unimpeded flow of
gradients.
\item A memory component is present in each layer in a serial manner where
the first layer sees $t-T$ time instant and the last layer sees $t-1$
frame. The component weights are shared across all layers to enable
them to learn longer-context.
\end{itemize}
A combination of these two components allows RMN to learn long-term
dependencies and higher level abstracts simultaneously in a much simpler
and efficient way. Bi-directional RMN (BRMN) is also formulated in
this work, which is a simple extension to RMN by adding an extra connection
with shared weights for learning future information. Computational
complexity is relatively less for BRMN over BLSTM or bi-directional
RNNs which is detailed in section \ref{sub:Bi-directional-residual-memory}.

In section \ref{sec:Experimental-setup}, we explain about AMI corpus
and the baseline model configurations used in our experiments. A detailed
explanation to build proposed RMN and BRMN model is in section \ref{sub:RMN-configuration-and}
and \ref{sub:BRMN-configuration-and}. Empirical evaluation is conducted
in section \ref{sec:Validation-experiments} to validate the structure
of RMN for speech recognition tasks. Comparison of the RMN with the
best LVCSR systems in literature in listed in section \ref{sub:Summary-of-results},
which is followed by conclusion and future work.

\section{Residual memory networks\label{sec:Residual-memory-networks}}

\begin{figure}[H]
\begin{centering}
\includegraphics[width=8.5cm,height=3.5cm]{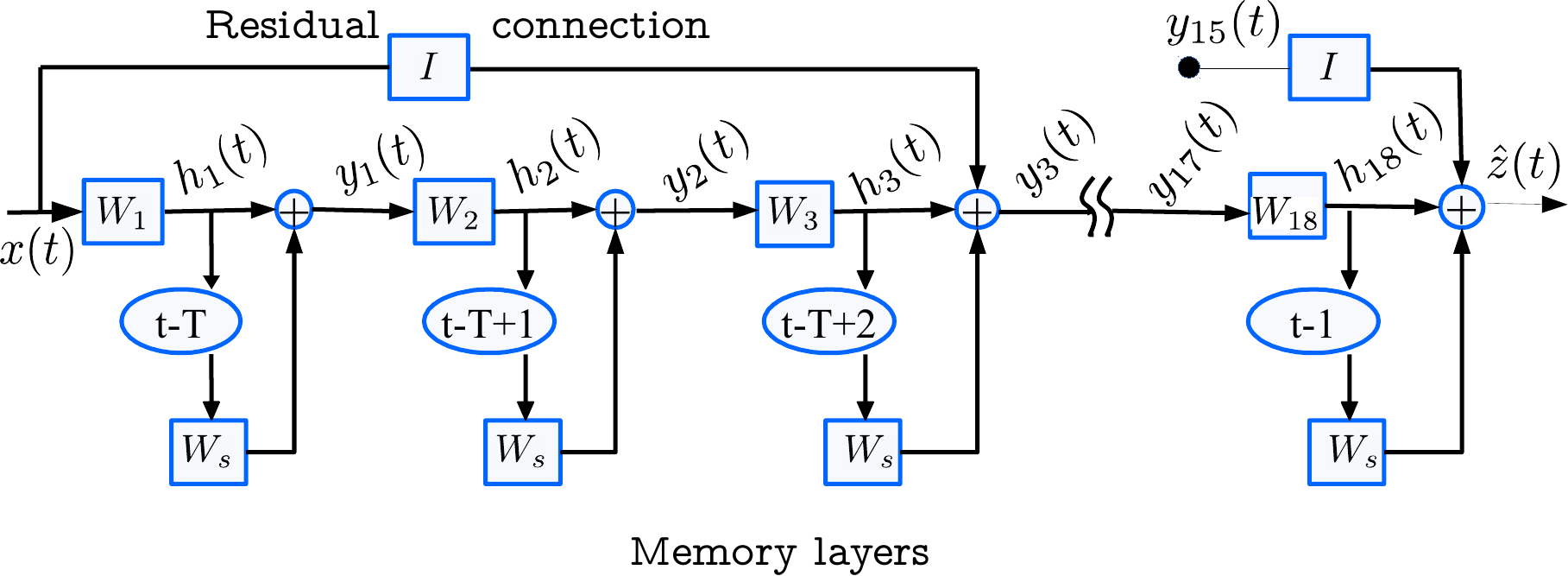}
\par\end{centering}

\caption{\label{fig:Architecture-of-residual}Architecture of residual memory
network (RMN) with number of memory layers $L=18.$ The memory layers
can model temporal context size of 18. }
\end{figure}

RMN is composed of memory layers and residual connections as shown
in figure \ref{fig:Architecture-of-residual}. The residual connection
connects the previous output to the current input by skipping few
layers. Each memory layer contains two weight transforms: The first
affine transform $W_{l}$ learns current time step and is different
for each layer $l=1,2,..L$ as in standard DNNs. The second weight
transform $W_{s}$ is shared across all layers and learns past information
by varying delay in decreasing order. For example in figure \ref{fig:Architecture-of-residual},
$W_{s}$ receives $t-T^{th}$ frame in first layer, second layer receives
$t-(T-1)^{th}$ frame and the delay keep decreasing as we proceed
to higher layers. In RMN $T$ is fixed based on the number of layers.
For instance 18 layered network captures 18 time steps. In this network,
relu activation is used after each memory layer as it is efficient
for training deeper networks \cite{ghahremani2016linearly,he2015deep}.
Thus, RMN can be represented as a variant of deep feed-forward neural
network which harnesses the important characteristics of unfolded-RNN
and residual networks.

\subsubsection{Forward propagation}

Figure \ref{fig:Architecture-of-residual} shows the series of computations
done in RMN architecture, where input $x(t),\,\{t=1,2,..T\}$ at time
instant $t$ is processed using $W_{l}$ matrix in the layer $l$
to get $h_{l}(t)$. The shared weight $W_{s}$ receives $h_{l}(t-m)$
by delaying $h_{l}(t)$ by $m$ time steps. The feed-forward output
after each memory layer is\vspace{-0.3cm}

\begin{equation}
y_{l}(t)=\phi\,(x(t)W_{l}+h_{l}(t-m)W_{s}),\,l=1,2,..L
\end{equation}

where $h_{l}(t)=x(t)W_{l}$ and $\phi$ is the relu activation output.

\subsubsection{Backward propagation}

Backpropagation for computing the parameter $W_{l}$ is done in the
same way as in standard DNNs. The shared parameter $W_{s}$ is computed
by taking into account error gradients from all $T$ time instants
which is exactly equal to $L$ memory layers. The error derivative
w.r.t to $W_{s}$ is\vspace{-0.3cm}

\begin{equation}
\frac{\partial E(t)}{\partial W_{s}}=\sideset{}{_{k=1}^{T}}\sum\left[\frac{\partial E(t)}{\partial\hat{z}(t)}\frac{\partial\hat{z}(t)}{\partial h_{l}(t)}\right]\frac{\partial h_{l}(t)}{\partial h_{k}(k)}\frac{\partial h_{k}(k)}{\partial W_{s}}
\end{equation}

where $\hat{z}(t)$ is the softmax output,$z(t)$ is target label
and $E_{t}(.)$ denotes cross-entropy loss function.

\subsection{Bi-directional residual memory network\label{sub:Bi-directional-residual-memory}}

In this section, the structure of bi-directional RMN (BRMN) is discussed.
The BRMN is an extension of RMN with one additional shared weight
transform which receives future frames as input. The forward propagation
output is given as\vspace{-0.3cm}

\begin{equation}
y_{l}(t)=\phi\,(x(t).W_{l}+h_{l}(t-m).W_{s}+h_{l}(t+m).W_{b})
\end{equation}
\vspace{-0.4cm}

where $t-m$ is the time instant delayed by m steps and $W_{b}$ is
the shared weight across layers. Unlike bi-directional RNN described
in \cite{schuster1997bidirectional}, BRMN does not require two separate
recurrent units for training future and past frames. The past and
future frames are not treated as independent entities and merged after
each layer. A possible explanation for bi-directional RNN to have
two separate layers is because RNN is tend to look over all frames
during prediction which leads to performance drop\cite{schuster1997bidirectional}.
In case of BRMN the network is constrained to pre-defined context
size based on the the number of memory layers and thus connecting
the forward states and backward states after each memory layer shows
improvement in performance. Also, BRMN requires only one extra weight
transform over RMN and hence the number of parameters is significantly
less when compared to bi-directional RNN and BLSTM \cite{graves2014towards,schuster1997bidirectional}.
\vspace{-0.5cm}

\section{Experimental setup\label{sec:Experimental-setup}}

The experiments were conducted on the AMI meeting conversation corpus
\footnote{http:corpus.amiproject.org/}, using the independent headset
microphone (IHM) recordings. The database is composed of 77 hours
- train data and 9 hours of each dev and eval data. 16KHz sampled
waveform was used to extract 13-dimensional MFCC features. These features
were mean normalized, spliced over 7 frames and projected down to
40 dimensions using linear discriminant analysis (LDA) obtained from
LDA+MLLT model. The LDA features were fed to speaker adaptive training
(SAT) using speaker based feature-space maximum likelihood linear
regression (fMLLR) transforms to obtain fMLLR features. 80 dimensional
log Mel-filterbank (fbank) features were also used for comparison.
The standard GMM-HMM and DNN is trained by following the Kaldi toolkit
\cite{povey2011kaldi}. The LSTM and RMN is trained using the CNTK
toolkit \cite{yu2014introduction}. SAT alignments using 4006 tied-states
were used as targets for neural network training. Testing was done
using eval set with trigram language model.

\subsection{Baseline DNN and LSTM models}

The DNN configuration includes 440 (40 x 11 splice) dimensional fMLLR
features at input and 4006 senones at softmax output. DNN containing
6 hidden layers, 2048 neurons were initialized using RBM pretraining
and fine-tuned with minibatch SGD frame-classification training. The
LSTM training is performed by following the CNTK recipe \cite{tan2016speaker}.
The LSTM is composed of 3 projected LSTM layers, each having 1024
memory cells and 512 projection units. The LSTM is trained using truncated
backpropagation through time (BPTT) with a minibatch size of 20. Forward
propagation is done with 40 parallel utterances for faster training
and better generalization \cite{yu2014introduction}. Highway LSTM
is also built by following the procedure in \cite{zhang2016highway}
to analyze the effect of increase in LSTM depth. 3 layered Highway
LSTM works better as increasing it to 8 degrades the performance as
in table \ref{tab:Baseline-recognition-performance}. Further experiments
in this paper is done with 3 layered Highway LSTM and will be denoted
as LSTM for simplicity. The BLSTM architecture is also used which
include 3 bi-directional layers each with 512 memory cells and 300
projection units. BLSTM is trained using latency control technique
with 22 past frames and 21 future frames as mentioned in \cite{zhang2016highway}.
The LSTM based models receives 40 dimensional fMLLR directly as splicing
does not help for training LSTMs \cite{miao2015speaker}. The speech
recognition results of all these models are stated in table \ref{tab:Baseline-recognition-performance}.

\subsection{Ivector extractor\label{sub:Ivector-extractor}}

BUT standardization initiative tool \footnote{http://voicebiometry.org/}
is used an ivector extractor. This extractor is trained on 9000 hours
of Fisher English (part 1 and 2), NIST SRE 2004-2008, Switchboard
(phase 2, phase 3, cellular part 1, and cellular part 2). An 100 dimensional
mean- and length-normalized ivector is extracted for each speaker,
after applying multi-lingual neural network based VAD tuned to detect
confident speech. Ivectors are appended to input features for training
LSTM based systems as suggested in \cite{tan2016speaker}.

\subsection{RMN configuration and training\label{sub:RMN-configuration-and}}

The memory layers in the RMN architecture contains 512 hidden units
followed by the relu activation function. The memory layer is preceded
and followed by a higher dimensional layer of 1024 units as it was
found to be crucial for better learning. Thus the network configuration
is represented as 440-1024- {[}512 x N layers{]}-1024-4006. The separate
weights of RMN $W_{l}$ as mentioned in section \ref{sec:Residual-memory-networks}
are Gaussian initialized with mean $0$ and standard deviation of
$0.2/\sqrt{layer\,dim}$. The shared weight matrix $W_{s}$ is initialized
as $0$ to ensure the model learns feature representation during the
first epoch followed by learning of temporal information. This is
done to disable the transfer of delayed inputs at the beginning of
training. Also, empirically we found that we can restrict the matrix
$W_{s}$ to be diagonal without loss in performance and it was used
throughout this work. In RMN, the residual connection shown is created
after every 3 layers as we found it to be optimum as in table \ref{tab:skips}.
Thus RMN requires smaller number of parameters compared to standard
DNN and LSTM as mentioned in table \ref{tab:param/wer-ami}. The
RMN is trained using truncated BPTT with a minibatch size of 256 as
suggested in \cite{saon2014unfolded} and a maximum of 10 utterances
in each minibatch. The initial learning rate is set to 0.2 and increased
to 1 in the next 4 epochs. Further training is done by automatically
reducing the learning rate for the next epoch by a factor of 0.5 if
cross-entropy loss degrades. L2 regularization weight is fixed to
0.00001 and momentum is set to 0.9. The scripts for running the RMN
and BRMN experiments are available in github repository \footnote{https://github.com/creatorscan/AMI\_CNTK\_scripts}

\subsection{BRMN configuration and training\label{sub:BRMN-configuration-and}}

The bi-directional RMN (BRMN) is trained by following the same procedure
as RMN with the following modifications: First, the initial learning
rate is fixed as 0.000095 and then allowed to auto adjust \cite{yu2014introduction}
based on validation loss as in RMN. Second, latency control technique
is used to capture 21 future frames. Empirically, we found that BRMN
works better for non-spliced input i.e., 40 dimensional fMLLR features
were directly sent to input of BRMN. Table \ref{tab:splice} shows
the performance BRMN with and without splicing. Thus the network configuration
of BRMN is 40-1024-{[}512 x 18 layers{]}-1024-4006. \vspace{-0.5cm}
\begin{table}[H]
\begin{minipage}[t]{1\columnwidth}%
\caption{\label{tab:Baseline-recognition-performance}Baseline recognition
performance (\% WER) of DNN, DNN+residual, LSTM, highway LSTM and
BLSTM for eval set of AMI corpus using fMLLR features}

\begin{center}
\vspace{-0.5cm}
\begin{tabular}{|c|c|c|c|c|c|}
\hline 
\multirow{2}{*}{DNN} & DNN+ & LSTM \cite{tan2016speaker} & \multicolumn{2}{c|}{Highway LSTM} & BLSTM\tabularnewline
\cline{4-6} 
 & residual & 3 layers & 3 layers & 8 layers & 3 layers\tabularnewline
\hline 
\hline 
27.1 & 26.7 & 26.0 & 25.8 & 26.0 & 24.9\tabularnewline
\hline 
\end{tabular}
\par\end{center}%
\end{minipage}\hfill{}%
\begin{minipage}[t]{1\columnwidth}%
\caption{\label{tab:skips}\% WER of RMN model with number of layers skipped
by the residual connection. This experiment is done by with 15 memory
layers of RMN}

\begin{center}
\vspace{-1cm}

\par\end{center}

\begin{center}
\begin{tabular}{|c|c|c|c|}
\hline 
\multicolumn{4}{|c|}{\# layers skipped by residue}\tabularnewline
\hline 
1 & 2 & 3 & 4\tabularnewline
\hline 
\hline 
26.6 & 26.0 & 25.9 & 25.9\tabularnewline
\hline 
\end{tabular}
\par\end{center}%
\end{minipage}\hfill{}%
\begin{minipage}[t]{1\columnwidth}%
\caption{\label{tab:splice}\% WER of RMN and BRMN for spliced and non-spliced
input with different width using eval set of AMI corpus}

\begin{center}
\vspace{-1cm}

\par\end{center}

\begin{center}
\begin{tabular}{|c|c|c|c|}
\hline 
\% WER & \multicolumn{3}{c|}{Splice width}\tabularnewline
\cline{2-4} 
Model & 1 (no splice) & 15 (+/-7) & 21 (+/-10)\tabularnewline
\hline 
\hline 
RMN & 26.6 & 25.6 & 25.6\tabularnewline
\hline 
BRMN & 24.3 & 24.8 & 25.4\tabularnewline
\hline 
\end{tabular}
\par\end{center}%
\end{minipage}
\end{table}
\vspace{-0.7cm}

\section{Validation experiments\label{sec:Validation-experiments}}

This section provides a detailed analysis to understand the behavior
of RMN.

\subsection{Effect on layer size and feature type}

\begin{figure}[H]
\includegraphics[scale=0.35]{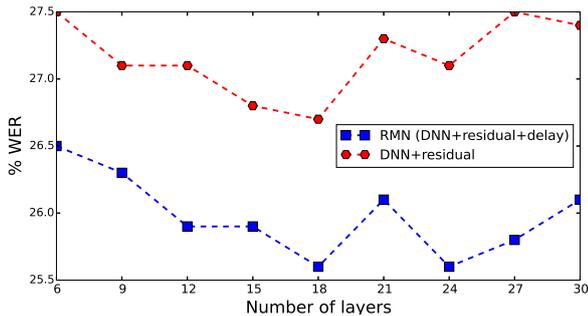}

\caption{\label{fig:res_rmn_dnn_layer_size}Performance comparison between
RMN and (DNN+residual) networks for different layer sizes}
\end{figure}

Initial experiment is done to find the optimum number of layers for
RMN using 40 dimensional fMLLR features. Figure \ref{fig:res_rmn_dnn_layer_size}
compares the performance of DNN + residual networks with and without
delay connections. The baseline performance of DNN+residual without
delay is shown in table \ref{tab:Baseline-recognition-performance}
Two notable observations can be made from this figure. 1) The importance
of delay connections is clearly visible from this figure as it shows
more than 1\% absolute gain over DNN+residual models. 2) The increase
in number of layers has positive influence over both DNN+residual
and RMN systems. The performance of both systems reaches threshold
after 18 layers. Based on these observations, the number of optimum
RMN layers is chosen to 18.

Table \ref{tab:features-fbank/fmllr} shows the effect of fMLLR features
and fbank features using RMN. This table shows that RMN shows comparable
performance to LSTM using fbank features. The performance of RMN is
clearly better for fMLLR features when compared to fbank features.
The LSTM models seems to perform slightly better over RMN systems
for fbank features. It is also visible that both RMN and LSTM suffers
with fbank due to the existence of strong correlations in time and
frequency. Thus, further experiments in this paper will be conducted
based on fMLLR features.

\subsection{Effect of bi-directional RMN and parameters}

Table \ref{tab:param/wer-ami} shows that the BRMN model gains absolute
1.3 \% improvement over RMN. Next, it is observed that RMN shows slight
improvement over LSTM and a similar pattern is noted in bi-directional
models. A striking difference between RMN and LSTM models is the number
of computational parameters. Even though the RMN has deeper structure
the number of parameters is greatly reduced. From table, it is visible
that RMN requires 28.9 \% lesser parameters than LSTM while BRMN needs
38.5 \% lesser parameters than BLSTM. The reason behind this parameter
difference in BRMN is due to its simple addition of one extra weight
transform for future frames over RMN model, while BLSTM requires two
separate LSTM layers for both directions.\vspace{-0.3cm}
 
\vspace{-0.5cm}

\begin{figure}[H]
\includegraphics[scale=0.35]{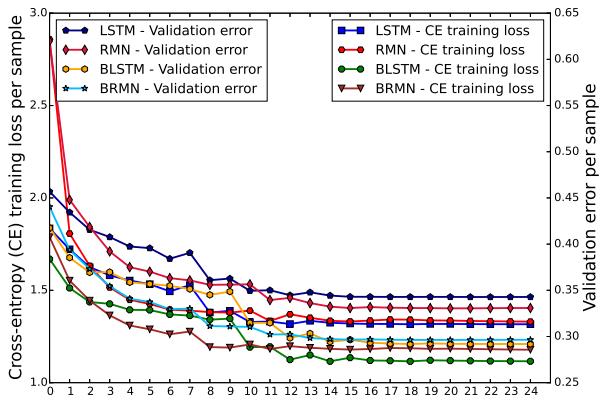}

\caption{\label{fig:loss}Training and validation err on RMN, BRMN, LSTM and
BLSTM using AMI corpus}
\end{figure}

To verify the importance of deeper architecture in reducing data mismatch,
the training and validation error is plotted in figure \ref{fig:loss}.
Convergence of RMN model indicates that RMN is learning in a similar
fashion as LSTM. The training loss of RMN models is more than when
compared to LSTM models, substantiating that RMN is less prone to
overfitting. In case of validation loss, RMN models shows less error
over LSTM based systems.\vspace{-0.3cm}
 \vspace{-0.7cm}

\subsection{Speaker adaptation using ivectors}

The RMN is fed with ivectors by augmenting it with input features.
The experiments in table \ref{tab:param/wer-ami} shows that RMN with
fMLLR+ivector features gives 6.25 \% relative improvement over RMN
with fMLLR. A relative improvement of 7.4 \% is obtained by BRMN model
using fMLLR+ivector over fMLLR features. A minimal improvement of
absolute 0.3 \% is obtained for RMN over LSTM network. The BRMN system
with ivectors provides 1.75 \% relative improvement over BLSTM with
ivectors. Augmenting ivectors in BLSTM increases the parameters by
2 million whereas BRMN parameters are increased by 0.1 million. This
signifies that BRMN shows significant gains over BLSTM without increasing
the parameters drastically.\vspace{-0.7cm}
\begin{table}[H]
\caption{Comparison of LSTM and RMN with uni- and bi-directional layers and
different types of features. Performance of speaker adaptation using
ivectors is noted here. The number of parameters (\# params) computed
is also listed. \label{tab:param/wer-ami}\label{tab:features-fbank/fmllr}}

\centering{}{\footnotesize{}}%
\begin{tabular}{|c|c|c|c|c|c|}
\hline 
\multirow{2}{*}{{\footnotesize{}Input}} & \multirow{2}{*}{{\footnotesize{}Model}} & \multicolumn{2}{c|}{{\footnotesize{}Uni-directional}} & \multicolumn{2}{c|}{{\footnotesize{}Bi-directional}}\tabularnewline
\cline{3-6} 
 &  & {\footnotesize{}\% WER} & {\footnotesize{}\# params} & {\footnotesize{}\% WER} & {\footnotesize{}\# params}\tabularnewline
\hline 
\hline 
\multirow{2}{*}{{\footnotesize{}Fbank}} & {\footnotesize{}LSTM } & {\footnotesize{}27.8} & {\footnotesize{}14.5 M} & {\footnotesize{}26.3} & {\footnotesize{}17.1 M}\tabularnewline
\cline{2-6} 
 & {\footnotesize{}RMN} & {\footnotesize{}28.1} & {\footnotesize{}10.7 M} & {\footnotesize{}26.4} & {\footnotesize{}9.9 M}\tabularnewline
\hline 
\multirow{2}{*}{{\footnotesize{}fMLLR}} & {\footnotesize{}LSTM } & {\footnotesize{}25.8} & {\footnotesize{}14.5 M} & {\footnotesize{}24.5} & {\footnotesize{}16.1 M}\tabularnewline
\cline{2-6} 
 & {\footnotesize{}RMN} & \textbf{\footnotesize{}25.6} & {\footnotesize{}10.3 M} & \textbf{\footnotesize{}24.3} & {\footnotesize{}9.9 M}\tabularnewline
\hline 
{\footnotesize{}fMLLR+} & {\footnotesize{}LSTM } & {\footnotesize{}24.3 } & {\footnotesize{}16.5 M} & {\footnotesize{}22.9} & {\footnotesize{}18.1 M}\tabularnewline
\cline{2-6} 
{\footnotesize{}ivec} & {\footnotesize{}RMN } & \textbf{\footnotesize{}24.0} & {\footnotesize{}10.4 M} & \textbf{\footnotesize{}22.5} & {\footnotesize{}10.0 M}\tabularnewline
\hline 
\end{tabular}{\footnotesize \par}
\end{table}
\vspace{-0.7cm}
\begin{table}[H]
\caption{\label{tab:stateoftheart}Comparison of RMN with existing methods
in literature trained using 300 hours of Switchboard corpus and tested
with Hub5-00 eval set. In this table 3g is trigram, 4g is meant as
4-gram, bn-fMLLR is bottleneck features with fMLLR and ivec represents
100 dimensional ivectors built using section \ref{sub:Ivector-extractor}.}

\centering{}{\footnotesize{}}%
\begin{tabular}{|c|c|c|c|c|}
\hline 
\multirow{2}{*}{\% WER} & \multirow{2}{*}{Model Type} & \multicolumn{3}{c|}{{\footnotesize{}SWB (\% CE WER)}}\tabularnewline
\cline{3-5} 
 &  & {\footnotesize{}3g} & {\footnotesize{}4g} & {\footnotesize{}4g+ivec}\tabularnewline
\hline 
\hline 
{\footnotesize{}Proposed } & {\footnotesize{}\hspace{-0.7cm}RMN} & {\footnotesize{}13.0} & {\footnotesize{}12.0} & \textbf{\footnotesize{}10.9}\tabularnewline
\cline{2-5} 
{\footnotesize{}Models} & {\footnotesize{}\hspace{-0.6cm}BRMN} & {\footnotesize{}11.8} & {\footnotesize{}10.8} & \textbf{\footnotesize{}9.9}\tabularnewline
\hline 
{\footnotesize{}State-of-the-art} & \multicolumn{3}{c|}{{\footnotesize{}\hspace{-2cm}TDNN \cite{povey2016purely}}} & {\footnotesize{}12.5}\tabularnewline
\cline{2-5} 
{\footnotesize{}results } & \multicolumn{3}{c|}{{\footnotesize{}Unfolded RNN + fMLLR \cite{saon2014unfolded}}} & {\footnotesize{}12.7}\tabularnewline
\cline{2-5} 
{\footnotesize{}in } & \multicolumn{3}{c|}{{\footnotesize{}\hspace{-0.5cm}LSTM + bn-fMLLR \cite{saon2016ibm}}} & {\footnotesize{}10.8}\tabularnewline
\cline{2-5} 
{\footnotesize{}literature} & \multicolumn{3}{c|}{{\footnotesize{}\hspace{-2.1cm}LSTM \cite{povey2016purely}}} & {\footnotesize{}11.6}\tabularnewline
\cline{2-5} 
 & \multicolumn{3}{c|}{{\footnotesize{}\hspace{-2cm}BLSTM \cite{povey2016purely}}} & {\footnotesize{}10.3}\tabularnewline
\hline 
\end{tabular}{\footnotesize \par}
\end{table}
\vspace{-0.5cm}

\subsection{Results of various LVCSR systems\label{sub:Summary-of-results}}

This section brings the state-of-the-results of LSTM, BLSTM, unfolded
RNN and TDNN models using 300 hours of switchboard data tested with
hub5-2000 evaluation set. These results are obtained by using ivectors
and 4-gram language model. To present a fair comparison, RMN model
is also trained with 300 hours data followed by testing with a trigram
language model built with Fisher and switchboard corpus, and later
rescored with 4-gram language model. The RMN and BRMN is trained in
a similar configuration as mentioned in section \ref{sub:RMN-configuration-and}
and \ref{sub:BRMN-configuration-and}. The table \ref{tab:stateoftheart}
shows the \% WER of switchboard (SWB) subset obtained after cross-entropy
(CE) training. The effectiveness of RMN in modeling long-term dependencies
for large vocabulary data is clearly visible in table \ref{tab:stateoftheart}.

\section{Conclusion}

In this paper we proposed residual memory network (RMN) structure,
a variant of feed-forward network to capture temporal context. We
have also introduced a bi-directional RMN which capture forward and
backward states with less computational complexity. The reasonable
effectiveness of RMN in capturing both temporal and higher-order information
was shown in AMI and switchboard tasks. BRMN showed 3.8 \% relative
improvement over BLSTM and RMN gained 6 \% relative improvement over
LSTM. In the future, we will develop a way to further increase the
depth of RMN for capturing longer context. A interesting direction
is to validate the ability of RMN in language modeling tasks. To increase
the efficiency of RMN with fbank features, we would like to use convolutional
layers before RMN to learn spatial representations.

\bibliographystyle{IEEEbib}
\bibliography{ref_new}

\end{document}